%% file: main.tex
\documentclass[a4paper,fleqn]{cas-dc}

\usepackage[authoryear]{natbib}
\usepackage[colorinlistoftodos]{todonotes}
\usepackage{longtable}

\begin{document}
\let\WriteBookmarks\relax
\def\floatpagepagefraction{1}
\def\textpagefraction{.001}

\shorttitle{US-JEPA}
\shortauthors{A. Radhachandran et~al.}

\title [mode = title]{US-JEPA: A Joint Embedding Predictive Architecture for Ultrasound}

\author[1,2]{Ashwath Radhachandran}[orcid=0000-0001-8793-0312]
\cormark[1]
\ead{ashwathradha123@g.ucla.edu}
\credit{Conceptualization, Data Curation, Methodology, Writing - review \& editing, Writing - original, Visualization, Formal analysis}

\author[1,6]{Vedrana Ivezić}
\credit{Conceptualization, Writing -- Review \& Editing}

\author[1,3]{Shreeram Athreya}
\credit{Conceptualization, Writing -- Review \& Editing}

\author[1,2,3,4,5,6]{Corey Arnold}
\credit{Conceptualization, Supervision, Resources, Writing -- Review \& Editing}

\author[1,2,5,6]{William Speier}
\credit{Conceptualization, Resources, Supervision, Funding acquisition, Writing -- Review \& Editing}

\affiliation[1]{organization={Biomedical Artificial Intelligence Research Lab, University of California, Los Angeles},
               city={Los Angeles},
               country={United States of America}}
\affiliation[2]{organization={Department of Bioengineering, University of California, Los Angeles},
               city={Los Angeles},
               country={United States of America}}
\affiliation[3]{organization={Department of Electrical and Computer Engineering, University of California, Los Angeles},
               city={Los Angeles},
               country={United States of America}}
\affiliation[4]{organization={Department of Pathology and Laboratory Medicine, University of California, Los Angeles},
               city={Los Angeles},
               country={United States of America}}
\affiliation[5]{organization={Department of Radiological Sciences, University of California, Los Angeles},
               city={Los Angeles},
               country={United States of America}}
\affiliation[6]{organization={Medical Informatics, University of California, Los Angeles},
               city={Los Angeles},
               country={United States of America}}

\cortext[cor1]{Corresponding author}

\begin{abstract}
Ultrasound (US) imaging poses unique challenges for representation learning due to its inherently noisy acquisition process. The low signal-to-noise ratio and stochastic speckle patterns hinder standard self-supervised learning methods relying on a pixel-level reconstruction objective. Joint-Embedding Predictive Architectures (JEPAs) address this drawback by predicting masked latent representations rather than raw pixels. However, standard approaches depend on hyperparameter-brittle and computationally expensive online teachers updated via exponential moving average. We propose US-JEPA, a self-supervised framework that adopts the Static-teacher Asymmetric Latent Training (SALT) objective. By using a frozen, domain-specific teacher to provide stable latent targets, US-JEPA decouples student--teacher optimization and pushes the student to expand upon the semantic priors of the teacher. In addition, we provide the first rigorous comparison of all publicly available state-of-the-art ultrasound foundation models on UltraBench, a public dataset benchmark spanning multiple organs and pathological conditions. Under linear probing for diverse classification tasks, US-JEPA achieves performance competitive with or superior to domain-specific and universal vision foundation model baselines. Our results demonstrate that masked latent prediction provides a stable and efficient path toward robust ultrasound representations.
\end{abstract}



\begin{keywords}
ultrasound \sep self-supervised learning \sep foundation model \sep JEPA
\end{keywords}

\maketitle


\section{Introduction}\label{sec:intro}

Self-supervised learning (SSL) has revolutionized the ability to leverage vast unlabeled datasets to build robust foundation models. In particular, masked image modeling (MIM) has emerged as a dominant approach, where models are tasked to reconstruct masked image patches to learn image semantics \citep{hondru_masked_2025}. Although pixel-level reconstruction objectives have proven effective in natural imaging, their utility partially depends on the assumption that local pixel intensities correlate strongly with underlying structural representations. However, when applied to domains characterized by low signal-to-noise ratio and acquisition artifacts, this assumption begins to break down \citep{xie_rethinking_2024}.

The ultrasound (US) imaging domain is a prime example of this representational bottleneck. US is a critical diagnostic tool in clinical settings, as it is portable, radiation-free, and provides real-time bedside feedback. Yet, image interpretation is difficult due to inherent visual graininess (speckle noise), operator dependence, and substantial variability in acquisition protocols and patient anatomy. In a traditional MIM framework, a generative, pixel-reconstruction objective may force the model to use its representational capacity to model these uninformative, acquisition-dependent features. Since these features, such as blur, acoustic shadow, and pixel-level contrast, vary significantly across clinical environments, a model trained for pixel reconstruction may overfit to these specific noise sources. The resulting significant gap between manually curated training data and real-world settings raises critical concerns about model robustness in out-of-distribution (OOD) scenarios. Without a pretraining objective that prioritizes semantic invariance, US foundation models may fail when subjected to diverse, yet realistic, ultrasound image corruptions.

The challenge of ensuring representational robustness is further compounded by the scarcity of high-quality clinical labels, which are expensive to acquire and require specialized medical expertise. Consequently, there is a need to train foundation models that can extract robust US feature representations and achieve high performance on downstream tasks using minimal supervision. Recent efforts to build US foundation models have attempted to navigate these bottlenecks through domain-specific modifications \citep{jiao_usfm_2024, megahed_usf-mae_2025, zhang_fully_2025}. Despite their innovations, these methods ultimately remain anchored to the pixel reconstruction paradigm, potentially underrepresenting the critical global features that distinguish different anatomical regions and pathologies in US.

\subsection{Ultrasound Joint Embedding Predictive Architecture (US-JEPA)}

To address these challenges, we introduce the Ultrasound Joint-Embedding Predictive Architecture (US-JEPA). Based on Image-based JEPA (I-JEPA), US-JEPA goes beyond the generative pixel-filling paradigm. Instead, it operates entirely in a latent embedding space, predicting the representations of masked target regions from a context block within the same image. By performing feature reconstruction rather than raw pixel reconstruction, US-JEPA can focus on learning global anatomical dependencies and tissue textures.

The I-JEPA framework \citep{assran_self-supervised_2023} relies on an online teacher updated via Exponential Moving Average (EMA) to guide a gradient-based student with smooth, yet improving targets. This coupling is computationally expensive and sensitive to hyperparameter selection. A core distinction of our framework is the adoption of the Static-teacher Asymmetric Latent Training (SALT) objective \citep{li_rethinking_2025}, in which a frozen teacher replaces an EMA-updated teacher. US-JEPA uses a frozen, domain-specific teacher, the Ultrasound Representation Foundation Model (URFM) \citep{kang_urfm_2025}, to provide stable targets.

URFM previously established the effectiveness of MIM for feature-level reconstruction in ultrasound, performing knowledge distillation from BiomedCLIP \citep{zhang_multimodal_2025}, a general-purpose medical vision encoder. URFM's reconstruction-based objective transfers rich semantic priors to the ultrasound domain, but remains tied to local reconstruction objectives as opposed to global representation consistency. By using URFM as a static teacher, US-JEPA takes a step further by leveraging the SALT latent prediction objective to refine URFM's ultrasound-specific representations. Rather than just reconstructing missing features, US-JEPA learns to align representations across views within an image. This approach encourages the model to learn anatomical context and global structure within US imaging, providing an improved latent space for downstream clinical tasks.

To ensure the model pretrained with US-JEPA is exposed to a wide range of human physiology, we aggregate one of the largest collections of publicly available US imaging to date. After thoroughly searching public repositories, our pretraining dataset encompasses approximately 5.12 million frames from 50 datasets covering 22 distinct anatomies.

\subsection{Standardizing Evaluation: UltraBench}

A significant challenge in current US foundation model research is the lack of a standardized evaluation protocol. Currently, most studies evaluate on disparate, often private, downstream datasets with non-standardized splits, making rigorous iteration and objective comparison nearly impossible. To promote reproducibility and establish a gold standard for future work, our paper adopts and extends UltraBench \citep{tupper_revisiting_nodate}, a publicly available benchmark for US frame-level evaluation. While the original UltraBench provides a strong foundation, we expand its scope to enhance anatomical diversity. Specifically, we include eight classification tasks spanning various organs and pathologies, adding two new tasks for thyroid and breast pathology.

Beyond the dataset benchmark itself, we address a critical gap in previous work: the lack of side-by-side comparisons of available models. To the best of our knowledge, this work is the first to perform an exhaustive linear probing evaluation across all published, publicly available US foundation models. Previous works compare against a limited set of US-specific baselines or rely solely on full fine-tuning, making it difficult to isolate the intrinsic quality of the learned representations. By tuning just a linear head for the classification tasks with a frozen backbone, we provide a standardized assessment of the learned latent space. This rigorous benchmarking ensures that improvements from US-JEPA are measured against a standard and relevant set of baselines, providing a clear map of the current state-of-the-art.

This work takes a significant step forward in ultrasound foundation modeling with the following contributions:
\begin{itemize}
\itemsep=0pt
\item \textbf{JEPA-based US Foundation Model:} We introduce US-JEPA, the first frame-level US foundation model built on JEPA principles.
\item \textbf{Label-Efficient Representations:} US-JEPA achieves strong linear probing performance with fewer labeled samples than competing baselines.
\item \textbf{Robustness to Domain-Specific Image Corruption:} Learned representations exhibit increased invariance to ultrasound-specific perturbations in image quality.
\item \textbf{Comprehensive Benchmarking:} We conduct the most comprehensive linear-probing evaluation to date across publicly available ultrasound foundation models on eight clinical tasks from UltraBench.
\item \textbf{Largest Public Pretraining Corpus:} We propose the OPUS (OPen UltraSound) dataset, the largest pretraining dataset to date for ultrasound composed exclusively from publicly available sources.
\end{itemize}

\section{Related Work}\label{sec:related}

\subsection{Universal Vision Foundation Models}

While early SSL focused on instance discrimination and contrastive objectives (e.g.\ MoCo \citep{he_momentum_2020} and SimCLR \citep{chen_simple_2020}), the field has expanded to self-distillation and predictive modeling to capture higher-level semantics. DINO \citep{caron_emerging_2021} demonstrated that student-teacher distillation with Vision Transformers (ViTs) can yield semantically rich features without labels. This paradigm has been scaled by successors, DINOv2 \citep{oquab_dinov2_2024} and DINOv3 \citep{simeoni_dinov3_2025}, to set new benchmarks in universal representation learning. In parallel, I-JEPA \citep{assran_self-supervised_2023} proposed moving beyond methods that predict in pixel or token space. By predicting missing latent representations from a visible context, I-JEPA learns the underlying structural logic of an image rather than superficial textures. Despite these universal models excelling on natural images, their direct application to US is limited by modality-specific artifacts, leading research to investigate specialized anatomical and task-based solutions.

\subsection{Task-Specific Ultrasound Foundation Models}

A significant branch of US research focuses on adapting universal architectures for specific clinical tasks, most notably through the specialization of the Segment Anything Model (SAM) \citep{kirillov_segment_2023}. Works such as SAMUS \citep{linguraru_beyond_2024} and UltraSAM \citep{meyer_ultrasam_2025} utilize architectural modifications or large-scale fine-tuning on large segmentation datasets to enable interactive, zero-shot segmentation. Beyond segmentation, SSL has proven useful for US representation learning within specific anatomical systems. In fetal US, UltraDINO \citep{ambsdorf_general_2025} outperforms general vision models and USFM by training DINOv2 from scratch on fetal-specific corpora. In the cardiac domain, EchoNet-Dynamic and EchoFM \citep{ouyang_video-based_2020, kim_echofm_2025} use large-scale US video corpora to learn temporal cardiac dynamics and apply these representations to clinical cardiac tasks. In parallel, others have pursued latent space prediction, with Ellis et al. introducing JEPAs to the domain by adopting V-JEPA for ultrasound video to prioritize temporal information over noisy pixels \citep{ellis_self-supervised_2025}. This approach was followed by EchoJEPA \citep{munim_echojepa_2026}, which demonstrated that JEPAs can develop a functional understanding of cardiac cycles in echocardiography. These successes demonstrate the power of domain-specific SSL; however, they often rely on privately curated, task-specific datasets, motivating the need for a robust, general-purpose US foundation model.

\subsection{General Ultrasound Foundation Models}

The evolution of US foundation models has been defined by a shift from standard pixel-level reconstruction toward domain-aware representation learning. USFM \citep{jiao_usfm_2024} was the first to define the field by training on more than 2 million images (public and private) using a dual spatial-frequency MIM objective, specifically designed to tackle the low spatial resolution and noise inherent in US. Building on this precedent, USF-MAE \citep{megahed_usf-mae_2025} demonstrated that pre-training exclusively on public US data (using the 370k-image OpenUS-46 corpus) combined with noise-filtering preprocessing could significantly surpass models initialized on natural images.

More recent efforts have integrated generative refinements and structural priors. For instance, D$^2$MAE \citep{kang_d2mae_nodate} uses diffusional deblurring in the MIM framework to address the low SNR inherent to US. EchoCare \citep{zhang_fully_2025} leverages a 4.5 million image dataset and introduces a hierarchical classification objective to teach the model nuanced anatomical relationships.

To build on these previous approaches, US-JEPA evolves the feature reconstruction paradigm introduced by URFM. By utilizing this domain-specific teacher and an asymmetric latent objective, US-JEPA is designed to encode the structural logic within US, learning an improved latent space for downstream clinical tasks.


\section{Preliminaries}\label{sec:prelim}

\subsection{I-JEPA}

The I-JEPA \citep{assran_self-supervised_2023} architecture consists of a context encoder $f_\theta$, a target encoder $\bar{f}_\theta$, and a predictor $g_\phi$. Given an input image $x$, the model operates on a sequence of non-overlapping patches. During training, we sample a context block $B_c$ and a set of $T$ target blocks $\{B_i\}_{i=1}^T$, where each block is a subset of patches. From these blocks, we derive disjoint masks $M_c$ and $M_i$ to ensure there is no overlap and prevent a non-trivial prediction task.

The target encoder processes the full image $x$ to produce patch-level representations $\mathbf{s}_y = \bar{f}_\theta(x)$. The targets for the loss function are defined as the subset of these representations corresponding to the $i$-th target mask, $\mathbf{y}_i = \mathbf{s}_y[M_i]$. Simultaneously, the context encoder processes only the visible image patches defined by $M_c$ to produce context embeddings $\mathbf{c} = f_\theta(x[M_c])$. The predictor $g_\phi$ then utilizes these context embeddings, conditioned on mask tokens $p_i$, which indicate the target-block locations. The mask tokens are parameterized by a learnable vector with an added positional embedding:
\begin{equation}
    \hat{\mathbf{y}}_i = g_\phi(\mathbf{c}, p_i)
\end{equation}

The training objective is to minimize the Smooth $L_1$ distance between the predicted $\hat{\mathbf{y}}_i$ and actual $\mathbf{y}_i$ target representations. To prevent representational collapse, I-JEPA employs an asymmetric update rule where the target encoder parameters $\bar{\theta}$ are an EMA of the context encoder parameters $\theta$.

\subsection{Static-teacher Asymmetric Latent Training (SALT)}

The SALT paradigm \citep{li_rethinking_2025} helps modulate the parameter selection in the JEPA objective by decoupling the student and teacher optimization. SALT demonstrates that a high-performing student can emerge by predicting the latent space of a static teacher, provided the teacher possesses reasonably sufficient semantic priors.

Here, the target encoder $h_\psi$ (e.g.\ an existing foundation model or encoder pretrained via a different objective such as pixel reconstruction) is frozen. The target embeddings $\mathbf{y}_i$ become static conditioned on the input image. Gradients propagate solely through the context encoder $f_\theta$ and predictor $g_\phi$:
\begin{equation}
    \mathcal{L}_{\text{SALT}} = \frac{1}{T} \sum_{i=1}^{T} \| g_\phi(f_\theta(x[M_c]), p_i) - h_\psi(x)[M_i] \|_1
\end{equation}
This approach eliminates the need for the EMA update, thus stabilizing the training dynamics and reducing computational overhead.

\section{US-JEPA}\label{sec:usjepa}

\begin{figure}
  \centering
  \includegraphics[width=1.0\columnwidth]{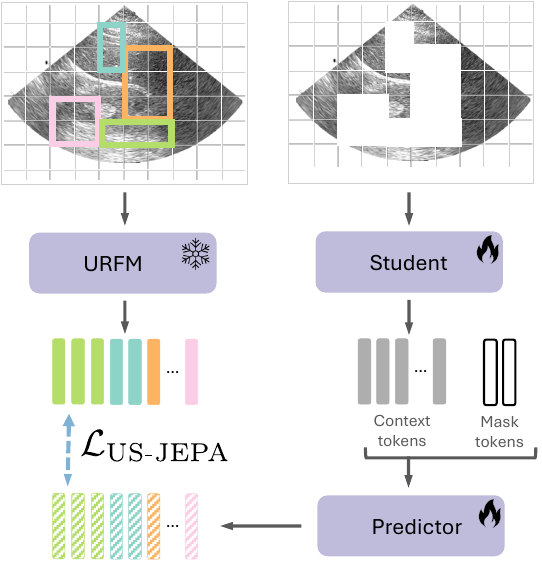}
  \caption{\textbf{USrc-JEPA framework.} URFM is the frozen teacher that extracts target embeddings. The student and predictor are jointly optimized with $\mathcal{L}_{\text{US-JEPA}}$ to align with the target.}
  \label{fig:model}
\end{figure}

US-JEPA (Fig.~\ref{fig:model}) relies on the masked latent objective from I-JEPA to learn the spatial semantics within ultrasound imaging. By forcing the student network to predict missing anatomical structures in a latent space with the guidance of a domain-specific teacher, the model is encouraged to develop a robust feature representation of tissue textures and organ morphology that is invariant to pixel-level noise.

\subsection{Self-Distillation via SALT}

While the SALT framework has shown that strong students can emerge even from sub-optimal teachers \citep{li_rethinking_2025}, our goal was to leverage the most robust US representations to establish a new state-of-the-art. To identify the optimal target provider, we performed an extensive benchmarking of existing US foundation models across UltraBench. Our preliminary results (detailed in Section~\ref{sec:experiments}) demonstrate that URFM consistently outperformed other baselines. We select URFM to serve as our frozen teacher and investigate whether the masked latent objective of US-JEPA can achieve better downstream performance, label efficiency in few-shot linear probing, and improved stability against domain-specific corruptions.

In our framework, the student context encoder and predictor are optimized to minimize the Smooth $L_1$ distance between the predicted target embeddings and the URFM target embeddings.

\subsection{Ultrasound Region-Conditioning (USrc)}

A significant challenge in training on diverse, public ultrasound datasets is the presence of non-anatomical artifacts. Ultrasound frames typically contain peripheral noise, including transducer metadata, intensity scales, patient information, and large black borders. Previous masked imaging approaches rely on raw frames, which is suboptimal since it allows non-ultrasound content to be sampled. We hypothesize that uniform random masking in I-JEPA might inadvertently task the model with predicting these meaningless regions, wasting representational capacity. To test this, we introduce USrc as a spatial prior to isolate the anatomical signal.

Let $x \in \mathbb{R}^{H \times W}$ be the input ultrasound image and $R \in \{0, 1\}^{H \times W}$ be a binary region mask, where $R_{ij}=1$ denotes valid ultrasound signal and $R_{ij}=0$ denotes background. We tile the image into non-overlapping patches and define the set of valid patches $\mathcal{P}_{valid}$ as those containing at least one pixel within $R$.

\textbf{Target Sampling:} Targets correspond to representations of image blocks. We feed $x$ through the frozen target encoder $h_\psi$ to obtain patch-level representations $\mathbf{s}_y = h_\psi(x)$. We sample $T$ candidate blocks $\{B_i\}_{i=1}^T$ based on scale and aspect ratio constraints. We employ a rejection sampling strategy where a candidate block is accepted only if its intersection with $\mathcal{P}_{valid}$ exceeds a threshold $\tau$:
\begin{equation}
    M_i = B_i \cap \mathcal{P}_{valid}, \quad \text{subject to } |M_i| \geq \tau
\end{equation}
The final targets are the representations $\mathbf{y}_i = \mathbf{s}_y[M_i]$.

\textbf{Context Sampling:} We sample a context block $B_c$ with specific scale and aspect ratio constraints. To ensure a non-trivial task, we remove any regions overlapping with the target blocks and apply our overlap constraint:
\begin{equation}
    M_c = B_c \cap \mathcal{P}_{valid}, \quad \text{subject to } |M_c| \geq \tau
\end{equation}
The masked context $x[M_c]$ is processed by the context encoder $f_\theta$ to obtain $\mathbf{c} = f_\theta(x[M_c])$.

\textbf{Optimization:} The training objective is minimized strictly over these anatomical intersections. The predictor aims to minimize the Smooth $L_1$ distance between the predicted student embeddings and the frozen teacher embeddings:
\begin{equation}
    \mathcal{L}_{\text{US-JEPA}} = \frac{1}{T} \sum_{i=1}^{T} \text{Smooth\,L1}\left( g_\phi(\mathbf{c}, p_{i}), \mathbf{y}_i \right)
\end{equation}

By constraining both input and output to $\mathcal{P}_{valid}$, US-JEPA with USrc (USrc-JEPA) engineers the inputs to only model tissue texture and anatomical structure.

\subsection{Training Pipeline}

To facilitate standardization across our aggregated corpus of US frames, we implement a standardized preprocessing and sampling pipeline for training.

\textbf{Preprocessing:} Images are first converted to grayscale. Regions with minimal colored artifacts or annotations are inpainted as long as they occupy less than 5\% of the total image area. The image is cropped to remove any negative space around the ultrasound content. To normalize for varying transducer gain settings across datasets, we perform intensity rescaling by mapping the $2^{nd}$ and $98^{th}$ percentiles of the pixel distribution to the full dynamic range. The USrc mask is used to condition the intensity rescaling only within the US content.

\textbf{Data Balancing Strategy:} A significant challenge in aggregating public medical datasets is the extreme variance in dataset size, which risks biasing the model toward the larger datasets. We mitigate this through a weighted dataset-level sampling strategy, where we define an upper bound sampling threshold $N_{t}$ of 50,000 to limit the effective size of each dataset per training epoch. For each dataset $D_i$ with actual size $|D_i|$, the effective count is $\min(|D_i|, N_{t})$, and the sampling probability is:
\begin{equation}
    P(D_i) = \frac{\min(|D_i|, N_{t})}{\sum_{j} \min(|D_j|, N_{t})}
\end{equation}

This strategy ensures that datasets larger than $N_{t}$ contribute equally during training, while smaller datasets contribute proportionally to their true size, balancing data diversity against over-representation of the largest sources.

\subsection{US-JEPA Architecture}

Both the student context encoder and the frozen URFM teacher employ a ViT-B/16 backbone with patch size 16 and embedding dimension 768, processing $224 \times 224$ inputs. The predictor is a narrower transformer with output dimension 384, followed by a linear projection head to align with the teacher's embedding space prior to Smooth $L_1$ loss computation. Per image, we sample one large context block (scale $[0.85, 1.0]$) and four target blocks (scale $[0.075, 0.125]$, aspect ratio $[0.75, 1.5]$), enforcing non-overlapping masks to ensure the prediction task is non-trivial. We train for 100 epochs with batch size 128 using AdamW. Full hyperparameters are provided in Appendix~\ref{app:D.1}.

\section{Experiments and Results}\label{sec:experiments}

\subsection{Data}\label{sec:data}

\subsubsection{Pretraining Dataset}

\begin{figure}[]
  \centering
  \includegraphics[width=1.0\columnwidth]{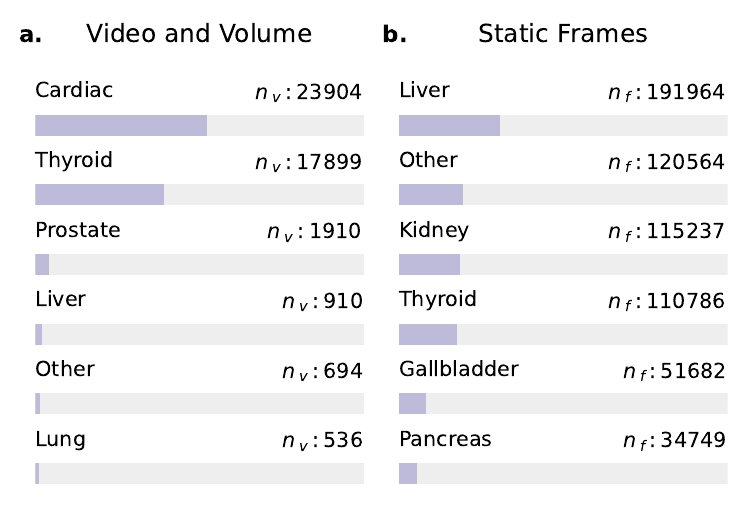}
  \caption{\textbf{Distribution of pretraining data.} To characterize the dataset composition at the organ level, we report the distribution of (a) temporal sequences, including videos and volumes ($n_v$), and (b) individual static frames ($n_f$).}
  \label{fig:data_dist}
\end{figure}

We introduce OPUS, to our knowledge the largest aggregation of publicly available ultrasound data, comprising 5,123,697 frames curated from 50 distinct publicly available datasets (Fig.~\ref{fig:data_dist}). Both US-JEPA and USrc-JEPA are pretrained on OPUS. The data contains three different US acquisition types, with a majority (79.7\%) of the frames derived from temporal video sequences and the remaining from volumetric (3D) scans or static 2D images. The dataset also spans a wide range of human anatomy, including major organ systems and important ancillary structures. Cardiac data represents the largest subset (23,904 videos), largely sourced from high-volume video datasets like EchoNet-Dynamic and EchoNet-LVH. This is complemented by significant representations of key organs, including the liver (318K frames), thyroid (264K frames), and prostate (227K frames), as well as breast (79K frames) and lung (55K). The data originates from a diverse array of clinical contexts, ranging from large-scale institutional repositories (e.g.\ Stanford, TCIA, RadImageNet) to more specialized, curated datasets, providing the model with exposure to varying image qualities, transducer frequencies, and pathological presentations. Further details regarding the OPUS pretraining dataset are included in Appendix~\ref{app:B.1}.

\subsubsection{UltraBench Downstream Dataset}

To ensure standardized and reproducible evaluation, we leverage UltraBench~\citep{tupper_revisiting_nodate}, a publicly available benchmark encompassing diverse ultrasound tasks across segmentation, detection, and classification. We add implementation for BUSBRA and TN5000 to UltraBench to expand anatomical coverage for breast and thyroid. Moreover, pretraining and downstream datasets are fully disjoint: the pretraining corpus OPUS does not include any of the eight UltraBench evaluation datasets. Our evaluation focuses on these eight classification tasks, comprising both binary and multi-class endpoints (Table~\ref{tab:downstream-datasets}), to assess the discriminative ability and representation quality of our models. Five of the eight datasets focus primarily on cancer detection. FATTY LIVER and POCUS have non-oncological applications in fatty liver disease detection and lung pathology detection (healthy, pneumonia, and COVID), respectively. Lastly, BUTTERFLY is a general multi-organ detection task. In Section~\ref{sec:fewshot}
and Section~\ref{sec:corruption},
our figures specifically show results for four datasets: FATTY LIVER and POCUS (non-cancer), MMOTU (challenging 8-class task), and BUSBRA (cancer-specific performance).

\begin{table}[width=.9\linewidth,cols=3,pos=h]
\caption{Summary of downstream datasets used for classification tasks, including the target organ, number of classes ($Y$), and total image counts. All datasets are public.}
\label{tab:downstream-datasets}
\begin{tabular*}{\tblwidth}{@{}LLR@{}}
\toprule
Dataset & Organ & Total Images \\
\midrule
AUL ($Y$=3)          & Liver       & 735   \\
BUSBRA ($Y$=2)       & Breast      & 1064  \\
BUTTERFLY ($Y$=9)    & Multi-Organ & 41076 \\
FATTY LIV.\ ($Y$=2) & Liver       & 550   \\
GBCU ($Y$=3)         & Gallbladder & 1255  \\
MMOTU ($Y$=8)        & Ovary       & 1469  \\
POCUS ($Y$=3)        & Lung        & 2064  \\
TN5000 ($Y$=2)       & Thyroid     & 5000  \\
\bottomrule
\end{tabular*}
\end{table}

\subsection{UltraBench Evaluation}\label{sec:ultrabench-eval}

To effectively evaluate the feature quality of US-JEPA and USrc-JEPA, we compare their performance against baselines using the standard SSL practice of linear probing \citep{alain_understanding_2018}. We identified six US foundation models with publicly available pretrained weights: USFM \citep{jiao_usfm_2024}, URFM \citep{kang_urfm_2025}, USF-MAE \citep{megahed_usf-mae_2025}, EchoCare \citep{zhang_fully_2025}, UltraSAM \citep{meyer_ultrasam_2025}, and SAMUS \citep{linguraru_beyond_2024}. UltraSAM and SAMUS are foundation models specifically developed for segmentation tasks, while the remaining are general ultrasound foundation models. Furthermore, we evaluate against two pretrained, state-of-the-art universal vision foundation models, DINOv3 and I-JEPA, which serve as out-of-domain benchmarks. To better isolate the contribution of our pretraining dataset, we use OPUS to train a ViT-Base under a standard MAE framework (MAE*) and retrain URFM from scratch (URFM*) using its defined architecture and hyperparameters. The final baseline is a ViT-Base trained with I-JEPA (I-JEPA*), using the original EMA-updated teacher setup to better understand the impact of SALT's frozen-teacher approach. Using the image encoder from each framework as a frozen feature extractor, we train linear probes for each downstream task across five random seeds to measure stability (Table~\ref{tab:results} and ~\ref{tab:results-controlled}).

\begin{table*}[t]
\caption{\textbf{Linear probe on downstream datasets.} We compare results of linear probes trained using our proposed method against state-of-the-art baselines. We report the mean macro F1 score \% over five seeds for all eight downstream datasets. Bold indicates the best result and underlining indicates the second best.}
\label{tab:results}
\begin{tabular*}{\tblwidth}{@{}LCCCCCCCC@{}}
\toprule
Model & AUL & BUSBRA & BUTTERFLY & FATTY LIV. & GBCU & MMOTU & POCUS & TN5000 \\
\midrule
DINOv3   & 64.3$\pm$0.6          & 70.9$\pm$1.7          & 91.7$\pm$0.4          & 55.8$\pm$5.5          & 61.7$\pm$0.5          & 37.2$\pm$0.6          & 91.4$\pm$0.4          & 67.5$\pm$0.4 \\
I-JEPA   & 61.5$\pm$1.1          & 71.2$\pm$4.0          & 90.5$\pm$0.6          & 54.8$\pm$1.6          & 53.7$\pm$0.4          & 35.3$\pm$0.6          & 88.1$\pm$0.4          & 68.9$\pm$0.2 \\
\midrule
UltraSAM & 62.6$\pm$3.1          & 70.2$\pm$3.1          & 89.6$\pm$2.4          & 66.9$\pm$3.3          & 43.5$\pm$4.9          & 39.7$\pm$1.8          & 87.3$\pm$2.1          & 63.9$\pm$2.0 \\
SAMUS    & 40.2$\pm$0.9          & 65.9$\pm$0.3          & 91.5$\pm$0.0          & 42.1$\pm$0.0          & 48.8$\pm$0.3          & 20.4$\pm$0.2          & 76.2$\pm$0.1          & 51.7$\pm$0.0 \\
EchoCare & 49.2$\pm$2.4          & 64.4$\pm$0.0          & 84.1$\pm$0.7          & 42.1$\pm$0.0          & 36.2$\pm$0.5          & 21.1$\pm$0.1          & 73.8$\pm$3.9          & 49.8$\pm$3.8 \\
USF-MAE  & 58.1$\pm$1.4          & 62.9$\pm$0.5          & 91.1$\pm$0.3          & 42.1$\pm$0.0          & 45.9$\pm$0.3          & 28.7$\pm$0.3          & 90.1$\pm$0.0          & 56.3$\pm$1.1 \\
USFM     & 61.6$\pm$1.2          & \underline{74.6$\pm$0.5} & \underline{92.4$\pm$0.3} & 73.6$\pm$8.8          & \underline{67.4$\pm$0.6} & 33.8$\pm$0.3          & 85.7$\pm$0.5          & 65.0$\pm$2.6 \\
URFM     & \underline{69.5$\pm$0.4} & 72.7$\pm$0.3          & \textbf{92.7$\pm$0.4} & 71.8$\pm$1.9          & 55.4$\pm$0.8          & 41.9$\pm$0.1          & 92.1$\pm$0.0          & \textbf{78.1$\pm$0.3} \\
\midrule
US-JEPA    & \textbf{69.6$\pm$1.5} & 73.8$\pm$1.1          & 90.8$\pm$0.3          & \underline{82.5$\pm$1.1} & 67.0$\pm$1.4          & \textbf{52.2$\pm$0.2} & \textbf{93.1$\pm$0.0} & \underline{73.1$\pm$0.7} \\
USrc-JEPA  & 67.6$\pm$0.5          & \textbf{76.0$\pm$1.2} & 91.5$\pm$0.6          & \textbf{89.2$\pm$0.9} & \textbf{70.2$\pm$0.5} & \underline{46.8$\pm$0.2} & \underline{92.5$\pm$0.1} & 70.8$\pm$1.3 \\
\bottomrule
\end{tabular*}
\end{table*}

\begin{table*}[t]
\caption{\textbf{Linear probe on downstream datasets — controlled pretraining comparison.} We compare our models against frameworks retrained on our pretraining corpus. MAE* denotes a ViT-Base trained under a standard masked autoencoder framework on our data; I-JEPA* denotes I-JEPA with an EMA-updated teacher trained on our data; URFM* denotes URFM retrained on our data using its original hyperparameters. We report the mean macro F1 score \% over five seeds for all eight downstream datasets. Bold indicates the best result and underlining indicates the second best.}
\label{tab:results-controlled}
\begin{tabular*}{\tblwidth}{@{}LCCCCCCCC@{}}
\toprule
Model & AUL & BUSBRA & BUTTERFLY & FATTY LIV. & GBCU & MMOTU & POCUS & TN5000 \\
\midrule
MAE*     & 64.5$\pm$0.0          & 70.4$\pm$0.5          & 89.0$\pm$0.2          & 82.4$\pm$0.0          & 62.5$\pm$0.0          & 41.0$\pm$0.4          & \underline{92.5$\pm$0.2} & 68.5$\pm$0.7 \\
I-JEPA*  & 57.8$\pm$1.3          & 69.2$\pm$1.3          & 82.8$\pm$0.6          & 83.6$\pm$7.2          & 52.2$\pm$1.2          & 29.8$\pm$0.2          & 83.8$\pm$0.5          & 55.5$\pm$3.4 \\
URFM*    & \textbf{72.1$\pm$0.2} & 71.2$\pm$0.2          & \textbf{93.6$\pm$0.1} & \textbf{89.9$\pm$0.0} & 62.9$\pm$0.2          & 45.9$\pm$0.5          & 90.9$\pm$0.0          & \underline{71.0$\pm$0.3} \\
\midrule
US-JEPA  & \underline{69.6$\pm$1.5} & \underline{73.8$\pm$1.1} & 90.8$\pm$0.3          & 82.5$\pm$1.1          & \underline{67.0$\pm$1.4} & \textbf{52.2$\pm$0.2} & \textbf{93.1$\pm$0.0} & \textbf{73.1$\pm$0.7} \\
USrc-JEPA & 67.6$\pm$0.5          & \textbf{76.0$\pm$1.2} & \underline{91.5$\pm$0.6} & \underline{89.2$\pm$0.9} & \textbf{70.2$\pm$0.5} & \underline{46.8$\pm$0.2} & \underline{92.5$\pm$0.1} & 70.8$\pm$1.3 \\
\bottomrule
\end{tabular*}
\end{table*}

\subsubsection{Direct Comparison}

US-JEPA and USrc-JEPA each achieve state-of-the-art performance on three of the eight tasks (AUL, MMOTU and POCUS for US-JEPA, and BUSBRA, FATTY LIV. and GBCU for USrc-JEPA). They each also fall in second place on two tasks (FATTY LIV. and TN5000 for US-JEPA, and MMOTU and POCUS for USrc-JEPA). BUTTERFLY is the only task that our models underperform against the top 2 baselines, URFM and USFM. However, even in this case, the difference is less than 2\% macro F1, and for a task where all models achieve competitive performance indicating it may be a relatively trivial task. Notably, our models excel in the most challenging scenarios where baseline methods struggle. The MMOTU dataset entails an eight-class ovarian tumor classification task. Average baseline performance drops below 40\% on MMOTU, while US-JEPA establishes a new benchmark of 52.2\%, surpassing the best baseline, URFM, by 10.3\%. While URFM performs strongly on TN5000, our models still offer superior performance on the remaining downstream tasks, and clearly beat general-purpose vision models (DINOv3 and I-JEPA) and the majority of the domain-specific baselines.

\subsubsection{Controlled Pretraining Comparison}

Table~\ref{tab:results-controlled} isolates the contributions of our pretraining corpus and architectural design by comparing baselines retrained on OPUS. MAE* and I-JEPA* underperform our models across all tasks, confirming that the pretraining data alone does not account for the gains from our framework. URFM*, which retrains the URFM architecture on OPUS, outperforms the original URFM on 5 of 8 downstreams, providing direct empirical support that OPUS yields stronger representations. The notable difference is with TN5000, where URFM achieves a macro F1 of 78.1 whereas URFM* drops to 71.0. Lastly, despite URFM* being a strong data-controlled baseline, US-JEPA outperforms URFM* on 5 of 8 tasks, demonstrating that our architectural contributions provide complementary gains beyond pretraining data alone.

\subsection{Few-Shot Scaling for Linear Probe}\label{sec:fewshot}

To evaluate the efficiency and transferability of our models' representations, we measure downstream performance with varying amounts of supervision. For each dataset, we train linear probes (across five seeds to measure stability) on every model's frozen features using stratified subsets of the training data: \{1\%, 5\%, 10\%, 50\%, 100\%\}. By preserving the original UltraBench splits and class prevalence through subsampling, we quantify how effectively pretrained features aid learning in low-shot regimes. Superior representation quality is evidenced by higher quantitative performance across all label densities, and by a less severe performance drop as label density decreases.

\begin{figure}
  \centering
  \includegraphics[width=\columnwidth]{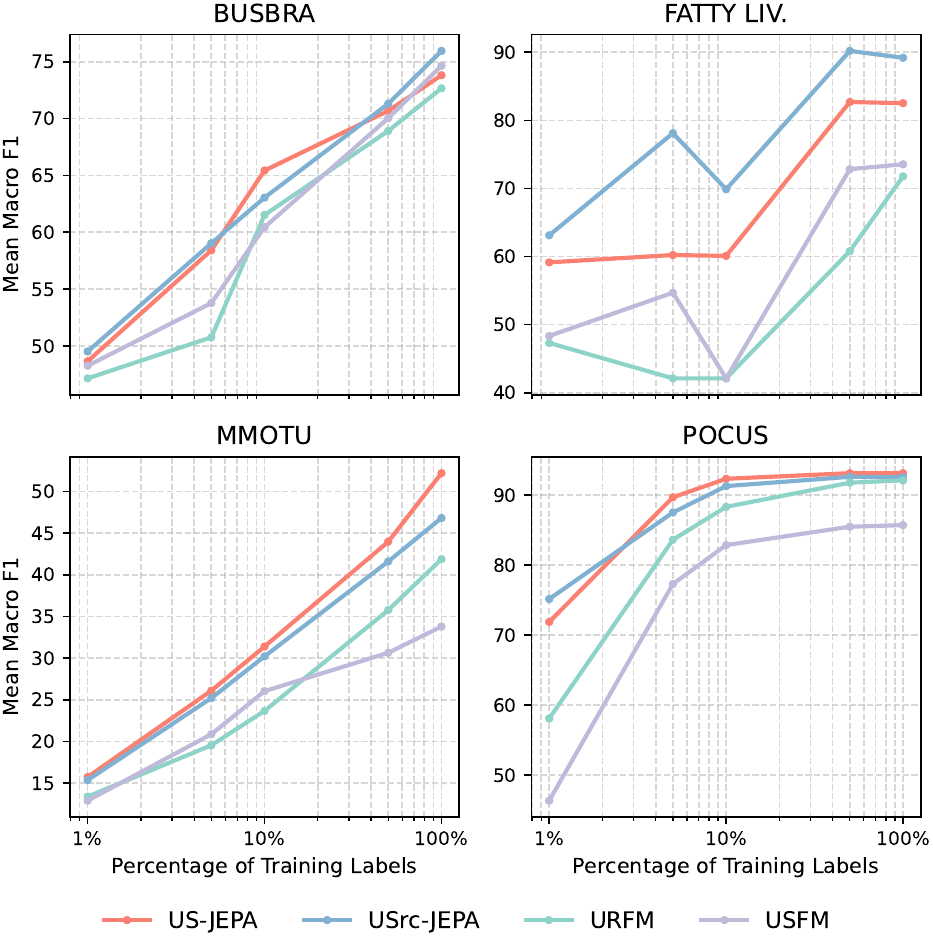}
  \caption{\textbf{Results for few-shot scaling.} We report the mean macro F1 score \% over five seeds for each model's probe to measure performance with 1\% to 100\% of training labels. \textit{Note: each dataset is plotted on a unique y-axis scale to better highlight model-specific performance trends.}}
  \label{fig:fewshot_2x2}
\end{figure}

Fig.~\ref{fig:fewshot_2x2} illustrates macro F1 performance across four datasets under varying label quantities for linear probing. We compare degradation against URFM and USFM. This pattern is most clearly seen with the FATTY LIVER downstream. When probes are trained with less than 50\% of labels, our models achieve an average macro F1 of 19\% higher than the URFM and USFM baselines. For the POCUS task, a similar trend is observed where the mean macro F1 degrades more severely for the baselines compared to our models. As probes are trained with fewer labels, even for BUSBRA, MMOTU, and GBCU (Appendix~\ref{app:E.1}), our models are at least on par, if not better than URFM and USFM. Complete results for all downstream tasks are included in Appendix~\ref{app:E.1}.

\begin{figure*}
  \centering
  \includegraphics[width=\textwidth]{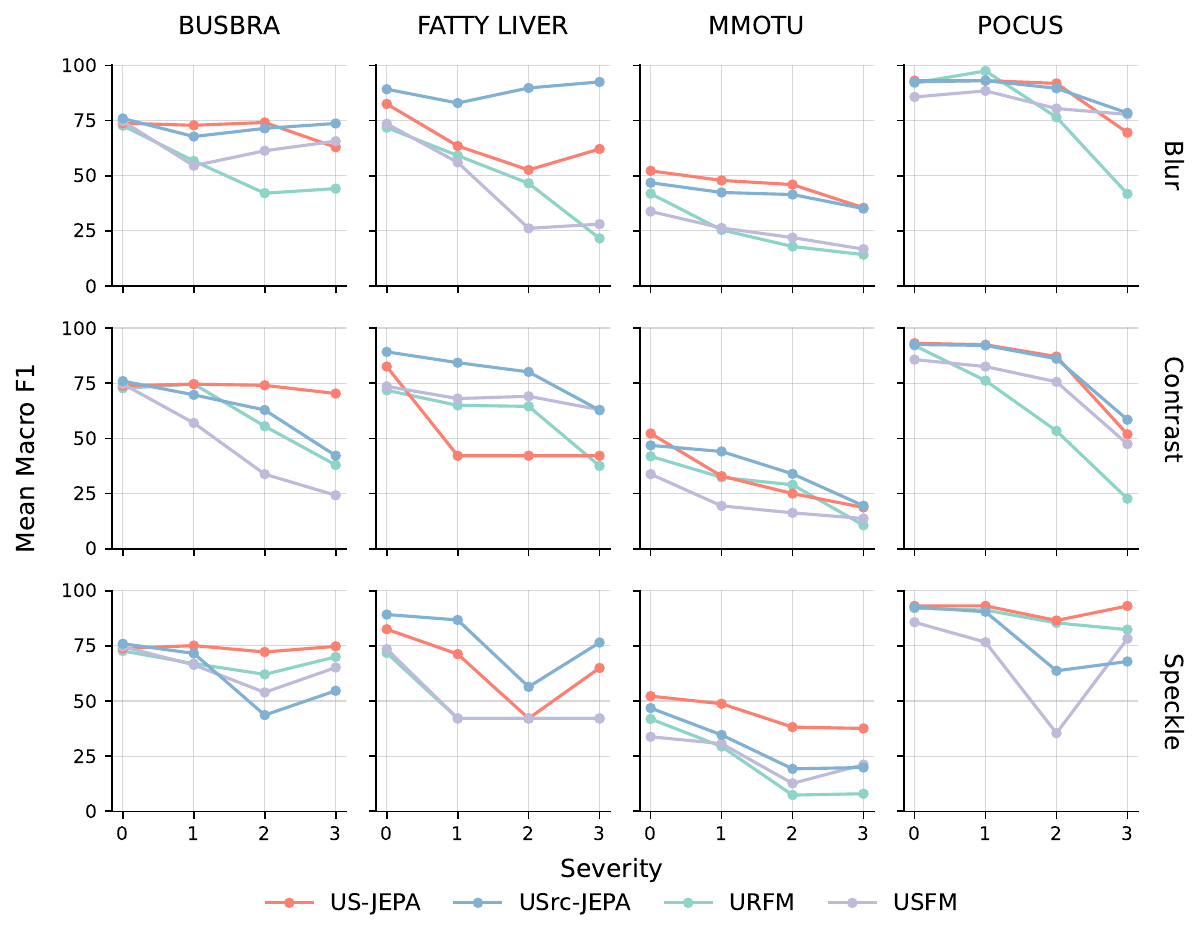}
  \caption{\textbf{Robustness to domain-specific corruption.} Results show the mean macro F1 across five seeds for each model--dataset--corruption permutation. Linear probes were trained on full, uncorrupted training sets and evaluated on increasingly corrupted test sets to assess structural representation stability.}
  \label{fig:oodrobustness_main}
\end{figure*}

\subsection{Robustness to Domain-Specific Corruption}\label{sec:corruption}

To evaluate the stability of US-JEPA and USrc-JEPA under distribution shifts, we perform an incremental stress test consisting of three synthetic, domain-specific image corruptions: Gaussian blur, contrast depletion, and correlated speckle noise. The first motivation for this is that in real-world clinical environments, ultrasound image quality is frequently compromised by hardware constraints, operator variability, and suboptimal acquisition settings. The second is to demonstrate that the JEPA-based pretraining objective encourages the model to learn global, structural relationships, and not expend model capacity on pixel-level variance.

For each corruption type, we evaluate downstream performance on test sets across three levels of increasing severity ($\epsilon \in \{1, 2, 3\}$). While blur and contrast are implemented via standard intensity transforms, we model speckle as a multiplicative, spatially-correlated process to approximate the ``grainy'' texture inherent to ultrasound scans. More detailed implementations for these corruptions are provided in Appendix~\ref{app:F.2}.

Fig.~\ref{fig:oodrobustness_main} visualizes the performance degradation for a subset of downstream tasks, with the remaining tasks detailed in Appendix~\ref{app:F.1}. Most notably, both US-JEPA and USrc-JEPA demonstrate significant resilience to the blur corruption. On the POCUS dataset, URFM degrades to less than half its uncorrupted performance, dropping from 92.1 to 41.8 F1 at maximum blur severity. In contrast, US-JEPA and USrc-JEPA are much more resilient, with F1 only dropping to 69.5 and 78.4 respectively at maximum blur severity. Similarly, on the BUTTERFLY dataset (Appendix~\ref{app:F.1}), URFM drops to 17.8 F1 under severe blur, while US-JEPA and USrc-JEPA maintain F1 performance at 80.8 and 70.9. In these severe blur cases, the second baseline USFM drops to 77.8 and 59.8 F1 for POCUS and BUTTERFLY, exhibiting less degradation than URFM but still performing worse than both our models for BUTTERFLY.

Regarding texture and intensity corruptions, our models match or exceed URFM and USFM for most downstream tasks. One unique weakness is observed on GBCU (Appendix~\ref{app:F.1}) where URFM maintains consistent performance with F1 only decreasing by 6.2\% from normal to the highest contrast severity, while our models drop by 47\%.


For speckle corruption, our models demonstrate remarkable stability. Specifically, on BUTTERFLY, US-JEPA and USrc-JEPA only drop by 0.6\% and 9.8\% under severe speckle noise ($\epsilon=3$), whereas USFM and URFM drop by 25\% and 44.1\%. Notably, across all downstream tasks US-JEPA outperforms or matches URFM and USFM under peak speckle severity (USrc-JEPA does so for 6/8 downstreams), demonstrating robustness to this prevalent ultrasound-specific noise. While USFM shows competitive results against USrc-JEPA on four tasks, it remains inferior to USrc-JEPA on the remaining, often by substantial margins, reaching a performance gap of up to 34.4\% on FATTY LIVER.

Ultimately, this stress test quantifies the invariance of our learned representations under significant corruption relative to the current state-of-the-art. By evaluating against domain-specific synthetic noise, we provide a practical validation of US-JEPA and USrc-JEPA's feature stability, a critical requirement given the high variance in US image quality across different manufacturers, operators, and clinical environments. The observed performance under corruption suggests that our approach is able to better capture structural semantics in US instead of overfitting to superficial, pixel-level features.

\subsection{Visualizing Learned Representations}

\begin{figure*}[]
  \centering
  \includegraphics[width=\textwidth]{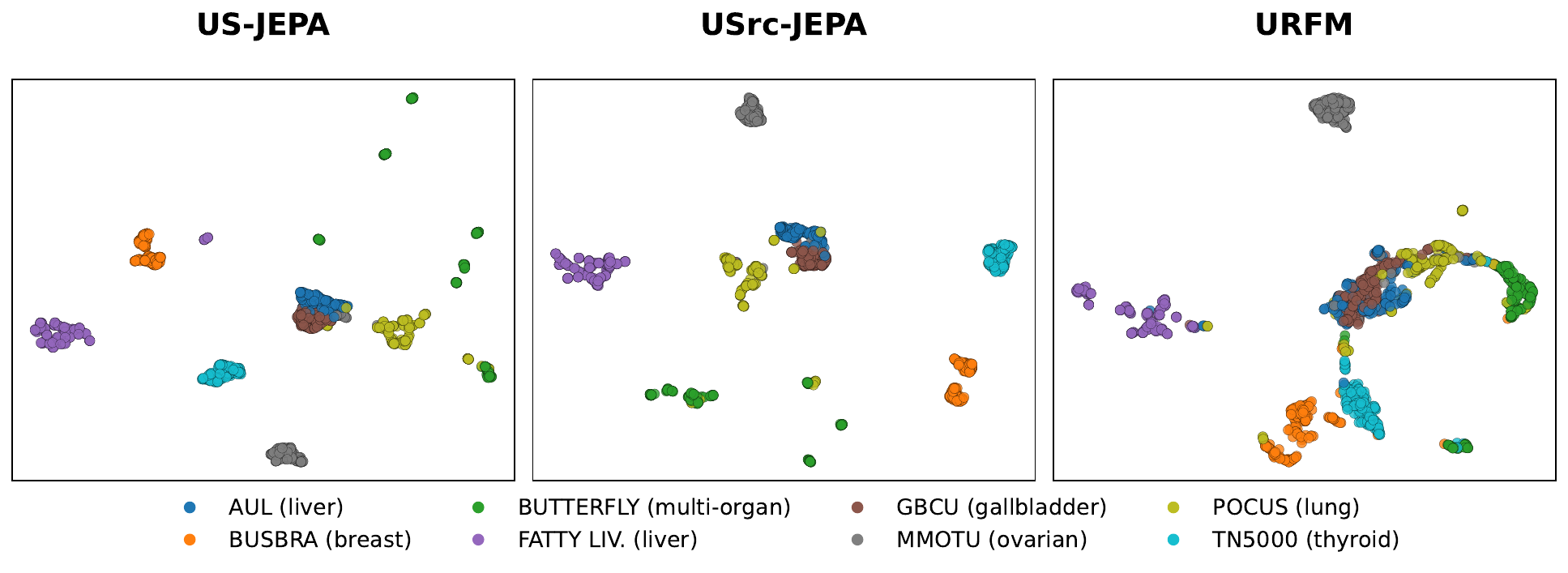}
  \caption{\textbf{UMAP feature visualizations.} UMAP projections of features extracted from US-JEPA, USrc-JEPA, and URFM across downstream datasets.}
  \label{fig:umap_comparison}
\end{figure*}

\textit{UMAP Feature Maps:} The US-JEPA and USrc-JEPA models produce generally distinct and separable clusters for each downstream dataset (Fig.~\ref{fig:umap_comparison}), whereas the URFM baseline demonstrates more overlap between representations for non-related anatomical structures (e.g., gallbladder and lung). We also note that our models adjacently cluster samples from the AUL (liver) and GBCU (gallbladder, with views of the liver) datasets, indicating that the features may capture overlapping anatomical regions. Lastly, for the multi-organ BUTTERFLY dataset, US-JEPA and USrc-JEPA form distinct clusters, which are correlated with their unique organ label.

\textit{Attention Maps:} Appendix~\ref{app:G} contains examples of attention maps to illustrate that our models consistently prioritize global structure and anatomical context. This trend is evidenced by specific observations across key datasets. In GBCU, our models capture liver and gallbladder boundaries along with surrounding peripheral tissue. In BUSBRA, our models attend to lesion contours, internal architecture, and stromal tissue. In AUL, our models maintain attention toward liver boundaries and broader tissue structures.

We observe similar patterns across the remaining datasets, where our models consistently capture the global structure and anatomy relevant to the task. These visualizations lend qualitative support to our claim that the JEPA objective encourages the model to learn the spatial continuity and relationships inherent to ultrasound rather than just focusing on clinically insignificant intensity differences or high-contrast regions.

\section{Discussion}\label{sec:discussion}

We present US-JEPA, a novel approach that demonstrates that training using a JEPA framework with a robust, static teacher yields stable, data-efficient representations for ultrasound. By adopting the SALT paradigm, we decouple the optimization of the student and teacher and leverage the semantic priors of a domain-specific teacher. Furthermore, with USrc-JEPA we show that adding USrc to the framework also achieves state-of-the-art performance on certain downstream tasks. This contribution is highlighted in our downstream stress tests: our models remain robust in low-data regimes for linear probing and maintain performance during evaluation with domain-specific corruptions. 


This work also advances standardization in ultrasound foundation model evaluation through three key contributions: 1) a rigorous linear probing comparison across all published models to establish standardized baselines for future benchmarking, 2) an expanded UltraBench incorporating TN5000 and BUSBRA for broader anatomical coverage and 3) the aggregation of OPUS, 
the largest publicly defined corpus of US imaging to date. By pretraining US-JEPA with OPUS, we demonstrate superior results against previous US foundation models on a majority of downstream tasks across relevant clinical domains.


\textit{Clinical Implications:} First, our investigation into learning label-efficient representations proved insightful for understanding which ultrasound foundation models are sample efficient. We observed that both US-JEPA and USrc-JEPA models were able to achieve as good, if not better, performance compared to the strongest baselines, USFM and URFM, in lower label regimes. These empirical results better position our proposed models to be more practically useful in resource-limited clinical settings, where annotated data may be difficult to acquire for fine-tuning, but a task-specific model is still necessary. Second, the experiments with domain-specific corruption provide a deeper understanding of how robust models are to synthetic, pixel-level corruptions. This experiment is an attempt to simulate real-world variability in visual appearance (e.g., transducer types, operators, etc.). More importantly, our models perform as well, or better, than baselines, USFM and URFM, under a majority of corruption-severity combinations.

\textit{Limitations:} Our robustness evaluation relies on synthetic corruptions (Gaussian blur, contrast depletion, and speckle) as reproducible proxies for clinical artifacts. While these stress tests assess a model's reliance on global structure over pixel-level texture, they do not fully capture complex ultrasound-specific phenomena such as acoustic shadowing or motion artifacts, whose occurrence depends on anatomical context and acquisition dynamics that are difficult to simulate without ground-truth maps. Future work will validate robustness against real-world distribution shifts, such as heterogeneous scanner settings, to more directly assess generalizability of models.

A second limitation concerns sensitivity to organ-level data density in pretraining. The performance gap between URFM and URFM* on thyroid nodule classification (TN5000), dropping from 78.1 to 71.0 macro F1, is likely attributable to the substantially lower thyroid prevalence in OPUS (5.2\%) relative to the original URFM pretraining corpus (44.2\%). This observation lends evidence to dataset scale alone being insufficient; anatomical or pathological balance in pretraining data composition may also meaningfully shape the quality of learned representations \citep{wang_data-_2026}. More broadly, underrepresented organs such as the thyroid and gallbladder may constitute failure modes for models pretrained on large-scale datasets without sufficient heterogeneity. Building a more principled understanding of how organ- and pathology-level density affects downstream representations remains an open challenge, and future work will extend our dataset-weighted sampling strategy to explicitly account for organ- and pathology-level prevalence during pretraining.

\textit{Future Directions:} Our work opens a new research direction in ultrasound foundation models by demonstrating that JEPA-based latent prediction is a viable and effective pretraining objective for this domain. While considerable effort has gone into frame-level foundation models, ultrasound video and volumetric imaging remain largely underexplored. Given the inherent spatiotemporal and volumetric structure in these modalities, masked latent reconstruction in the representation space offers a well-suited inductive bias for learning meaningful features. Beyond pretraining, broader task-specific downstream evaluation, spanning segmentation, detection, and regression alongside classification, would help surface application areas where current models struggle. And lastly, more systematic comparisons with task- and anatomy-specific foundation models would clarify how pretraining data composition and architectural design jointly shape learned representations and downstream performance.

Our findings confirm the utility of the JEPA framework for ultrasound self-supervised learning and provide an argument that joint-embedding predictive architectures could be a powerful way forward in learning stronger features from ultrasound imaging.

\printcredits

\section*{Declaration of generative AI and AI-assisted technologies in the manuscript preparation process}

During the preparation of this work the authors used Large Language Models (LLMs) such as ChatGPT and Gemini in order to assist with structural editing and formatting of the manuscript. The majority of this manuscript was not written with the help of a Large Language Model and assistance was limited to these structural and formatting tasks. After using these tools, the authors reviewed and edited the content as needed and take full responsibility for the content of the published article.

\section*{Funding}

This work was supported by the UCLA Radiology Exploratory Research Grant.

\onecolumn
\appendix
\counterwithin{figure}{section}
\counterwithin{table}{section}
\counterwithin{equation}{section}
\input{appendix}






\bibliographystyle{cas-model2-names}

\bibliography{references}


\end{document}

%% file: appendix.tex
\section{Dataset Details}\label{app:A}

\subsection{OPUS: Dataset Composition}\label{app:B.1}
For complete transparency, we have included the name of each dataset and associated metadata in OPUS.

\begin{longtable}{lllr}
\caption{Comprehensive breakdown of datasets used, categorized by anatomical region, frame type, and total frame count.} \label{tab:dataset_breakdown} \\
\toprule
\textbf{Dataset Name} & \textbf{Anatomy} & \textbf{Frame Type} & \textbf{Frame Count} \\ \midrule
\endfirsthead
\toprule
\textbf{Dataset Name} & \textbf{Anatomy} & \textbf{Frame Type} & \textbf{Frame Count} \\ \midrule
\endhead
\bottomrule
\endfoot
ultrasoundcases.info & Abdominal & Static & 3199 \\
abdominal\_US \citep{vitale_improving_2020} & Abdominal & Static & 617 \\
ultrasoundcases.info & Appendix & Static & 781 \\
ultrasoundcases.info & Bladder & Static & 450 \\
USAnotAI \citep{kim-ann_ftsvdusanotai_2024} & Bladder & Static & 61 \\
RadImageNet \citep{mei_radimagenet_2022} & Bladder & Static & 758 \\
USAnotAI & Bowel & Static & 61 \\
ultrasoundcases.info & Brain & Static & 466 \\
ReMIND \citep{juvekar_brain_2023} & Brain & Video & 18734 \\
ultrasoundcases.info & Breast & Static & 4956 \\
BUSI \citep{al-dhabyani_dataset_2019} & Breast & Static & 780 \\
BUS\_UC \citep{iqbal_bus_uc_2023} & Breast & Static & 811 \\
Breast-Lesions-USG \citep{pawlowska_curated_2024} & Breast & Static & 256 \\
BUSC \citep{iqbal_busc_2023} & Breast & Static & 250 \\
DatasetA \citep{jimenez_ultrasound_2024} & Breast & Static & 250 \\
BUV \citep{lin_new_2022} & Breast & Video & 25026 \\
BUID \citep{abbasian_ardakani_open-access_2023} & Breast & Static & 205 \\
ABUS-TDSC \citep{luo_tumor_2025} & Breast & Volume & 44474 \\
Breast\_S1 \citep{guo_segmentation_2021} & Breast & Static & 201 \\
UDIAT \citep{yap_automated_2018} & Breast & Static & 163 \\
BUSI\_WHU \citep{huang_busi_whu_2023} & Breast & Static & 927 \\
BUS\_UCLM \citep{vallez_bus-uclm_2024} & Breast & Static & 683 \\
cardiacCCAUS \citep{agata_momot_common_2022} & Cardiac & Static & 1100 \\
ultrasoundcases.info & Cardiac & Static & 1443 \\
EchoNet-LVH \citep{duffy_high-throughput_2022} & Cardiac & Video & 1957928 \\
Carotid US Boundary Study \citep{kristen_meiburger_dataset_2022} & Cardiac & Static & 500 \\
CAMUS \citep{leclerc_deep_2019} & Cardiac & Video & 19232 \\
cardiacUDC \citep{yang_graphecho_2023} & Cardiac & Video & 38071 \\
EchoNet: Tee-View-Classifier \citep{steffner_deep_2024} & Cardiac & Video & 55261 \\
EchoNet Dynamic \citep{ouyang_echonet-dynamic_nodate} & Cardiac & Video & 1770636 \\
RadImageNet & Cardiac & Static & 19062 \\
RadImageNet & Fibroid & Static & 2095 \\
GB1 \citep{turki_gallblader_2025} & Gallbladder & Static & 10692 \\
USAnotAI & Gallbladder & Static & 61 \\
RadImageNet & Gallbladder & Static & 39743 \\
ultrasoundcases.info & Gallbladder & Static & 1186 \\
GIST\_EUS \citep{he_query2mathmsup_2023} & Gastrointestinal & Static & 514 \\
ultrasoundcases.info & Gastrointestinal & Static & 2618 \\
USAnotAI & Kidney & Static & 61 \\
TRUSTED \citep{ndzimbong_trusted_2025} & Kidney & Volume & 7280 \\
OpenKidney \citep{singla_open_2022} & Kidney & Static & 500 \\
RadImageNet & Kidney & Static & 111252 \\
ultrasoundcases.info & Kidney & Static & 3924 \\
RadImageNet & Liver & Static & 78535 \\
ultrasoundcases.info & Liver & Static & 2714 \\
USAnotAI & Liver & Static & 61 \\
SYSU-CEUS-FLL \citep{liang_recognizing_2014} & Liver & Sampled Video & 110654 \\
B-mode-and-CEUS-Liver \citep{eisenbrey_ultrasound_2021} & Liver & Video & 126543 \\
COVID-Blues \citep{Born2021AcceleratingAnalysis} & Lung & Video & 31696 \\
COVIDx-US \citep{ebadi_covidx-us_2022} & Lung & Video & 22697 \\
ultrasoundcases.info & Lung & Static & 714 \\
ultrasoundcases.info & Lymph node & Static & 320 \\
ultrasoundcases.info & Muscle & Static & 17459 \\
Mus-V \citep{geng_force_2024} & Muscle & Sampled Video & 3114 \\
MuscleUS \citep{meiburger_dataset_2021} & Muscle & Static & 8169 \\
UBPD \citep{ding_mallesnet_2022} & Nerve & Sampled Video & 955 \\
ultrasoundcases.info & Nerve & Static & 458 \\
NerveUS \citep{ultrasound-nerve-segmentation} & Nerve & Static & 11143 \\
ovarianUS \citep{borna_ai-powered_2025} & Ovarian & Static & 301 \\
PCOS\_US \citep{choudhari_pcos_nodate} & Ovarian & Static & 3841 \\
RadImageNet & Ovarian & Static & 3595 \\
ultrasoundcases.info & Pancreas & Static & 1506 \\
LEPset \citep{li_lepset_2023} & Pancreas & Static & 11493 \\
105US \citep{hann_algorithm_2017} & Pancreas & Static & 105 \\
RadImageNet & Pancreas & Static & 21645 \\
RadImageNet & Portal Vein & Static & 769 \\
ProstateTRUS \citep{baum_mr_2023} & Prostate & Volume & 5037 \\
Prostate-MRI-US-Biopsy \citep{natarajan_prostate_2020} & Prostate & Volume & 219484 \\
MUPSD \citep{shao_micro-ultrasound_2024} & Prostate & Volume & 2910 \\
ultrasoundcases.info & Reproductive & Static & 3446 \\
RadImageNet & Spleen & Static & 6520 \\
ultrasoundcases.info & Spleen & Static & 1207 \\
USAnotAI & Spleen & Static & 61 \\
RadImageNet & Thyroid & Static & 92598 \\
SegThy \citep{kronke_tracked_2022} & Thyroid & Volume & 136243 \\
ultrasoundcases.info & Thyroid & Static & 2011 \\
Thyd2 \citep{hou_ultrasonography_2024} & Thyroid & Static & 8489 \\
DDTI \citep{pedraza_open_2015} & Thyroid & Static & 610 \\
TG3K \citep{gong_haifangongtrfe-net-for-thyroid-nodule-segmentation_2026} & Thyroid & Static & 3585 \\
TN3K \citep{gong_haifangongtrfe-net-for-thyroid-nodule-segmentation_2026} & Thyroid & Static & 3493 \\
Stanford Thyroid Cine \citep{yamashita_toward_2022} & Thyroid & Video & 17412 \\
RadImageNet & Uterus & Static & 13312 \\
Yang \citep{tiantian_yang_uterine_2023} & Uterus & Static & 1973 \\
\end{longtable}

\subsection{Downstream Datasets}\label{app:B.2}

Below, we provide a more detailed description of each downstream dataset, summarized in Table~\ref{dataset-detailed-table}. We use the publicly available UltraBench tool to set up the datasets and corresponding training, validation and test splits. We contribute two more datasets, BUSBRA and TN5000, to UltraBench for a more comprehensive downstream evaluation.

\textbf{Annotated Ultrasound Liver (AUL):} The Annotated Ultrasound Liver (AUL) dataset is a collection of 2D liver ultrasound images introduced for research on automated liver lesion analysis and mass characterization~\citep{Xu2023ImprovingFrames}. It comprises images acquired from distinct patients and includes expert annotations delineating the liver boundary and, when present, lesion contours. Each image is additionally assigned a clinical label reflecting the presence and type of focal liver mass, enabling both region-aware and image-level learning. The dataset spans a wide range of image resolutions and visual appearances, capturing realistic variability in ultrasound acquisition and anatomy. We use AUL for a supervised image-level mass classification downstream task with three classes: malignant mass, benign mass, and normal (no mass). The full dataset contains 735 images in total. We follow a fixed split consisting of 529 images for training, 59 for validation, and 147 for testing, ensuring that all splits remain patient-independent and preserve the original class distribution.

\textbf{Butterfly:} The Butterfly dataset was released by Butterfly Network for the 2018 MIT Grand Hack to promote research in point-of-care ultrasound understanding~\citep{ButterflyNetwork}. It comprises ultrasound images acquired using the Butterfly iQ device from multiple anatomical regions across 31 patients, reflecting the diversity and variability encountered in real-world bedside imaging. Each image is labeled according to the anatomical site being examined, enabling supervised learning for anatomical recognition in ultrasound. We use the Butterfly dataset for a supervised organ classification downstream task with nine classes: Morison’s pouch, bladder, heart (parasternal long-axis view), heart (four-chamber view), heart (two-chamber view), inferior vena cava, carotid artery, lungs, and thyroid. The dataset contains a total of 41,076 images, with 28,053 images used for training, 6,272 for validation, and 6,751 for testing. All splits are patient-independent to prevent data leakage and ensure fair evaluation.

\textbf{Fatty Liver:} The Fatty Liver dataset consists of B-mode liver ultrasound images introduced for studying automated detection of hepatic steatosis and non-alcoholic fatty liver disease (NAFLD)~\citep{Byra2018TransferImages}. The dataset includes images acquired from multiple patients with expert clinical labeling based on the presence of fatty infiltration in the liver parenchyma. It has been widely used to evaluate representation learning and transfer learning methods for liver ultrasound analysis, particularly in low-data clinical settings. We use this dataset for a supervised image-level fatty liver discrimination downstream task with two classes: normal liver and NAFLD. The dataset contains a total of 550 images, drawn from 55 patients. We use 390 images for training, 50 for validation, and 110 for testing. All splits are patient-independent to ensure a fair assessment of generalization performance.

\textbf{GBCU:} The Gallbladder Cancer Ultrasound (GBCU) dataset was introduced to support research on automated gallbladder lesion characterization from abdominal ultrasound images~\citep{Basu2022SurpassingLearning}. It consists of expert-annotated images collected from a diverse patient cohort and labeled according to underlying pathology, enabling supervised learning for clinically relevant gallbladder cancer assessment. The dataset captures a range of normal and diseased presentations and has been used to benchmark deep learning methods for malignancy discrimination in ultrasound. We use GBCU for a supervised image-level lesion malignancy classification downstream task with three classes: normal, benign, and malignant. The full dataset contains 1,255 images in total. We use 1,019 images for training, 114 for validation, and 122 for testing. The original dataset provides a fixed train–test split with no patient overlap; the validation set is created by further splitting the training data, as patient identifiers are not available to enforce strict separation.

\textbf{MMOTU:} The Multi-Modality Ovarian Tumor Ultrasound (MMOTU) dataset was introduced to facilitate research on ovarian tumor analysis using ultrasound and contrast-enhanced ultrasound imaging~\citep{zhao2023mmotu}. It provides expert annotations and diagnostic labels for a diverse set of ovarian pathologies, supporting both localization and classification tasks. In this work, we consider only the 2D ultrasound images and associated labels, focusing on image-level learning for ovarian tumor characterization. We use MMOTU for a supervised multi-class image-level tumor classification downstream task with eight classes: chocolate cyst, serous cystadenoma, teratoma, theca cell tumor, simple cyst, normal ovary, mucinous cystadenoma, and high-grade serous cystadenocarcinoma. The dataset contains 1,469 images in total. We use 800 images for training, 200 for validation, and 469 for testing. The original dataset provides a fixed train–test split, and the validation set is obtained by further splitting the training data, as patient identifiers are not available to enforce strict patient-level separation.

\textbf{POCUS:} The Point-of-Care Ultrasound (POCUS) dataset was introduced to support automated diagnosis of lung pathologies, particularly COVID-19, from bedside ultrasound imaging~\citep{Born2021AcceleratingAnalysis}. It aggregates lung ultrasound data acquired using convex and linear probes from multiple sources, reflecting real-world variability in point-of-care acquisition. The dataset has been widely used to benchmark deep learning methods for pulmonary disease recognition in ultrasound under heterogeneous data conditions. We use POCUS for a supervised image-level lung pathology classification downstream task with three classes: healthy, pneumonia, and COVID-19. Following the preprocessing protocol described in the original work, we use 29 convex-probe images and extract frame-level samples from 124 convex-probe videos while grouping frames by video to avoid data leakage across splits. The resulting dataset contains 2,064 images in total, with 1,444 images used for training, 177 for validation, and 443 for testing, ensuring strict separation at the video level.

\textbf{BUSBRA:} The BUS-BRA (Breast Ultrasound Brazil) dataset was introduced to support the development and standardized evaluation of computer-aided diagnosis systems for breast ultrasound imaging, with an emphasis on clinically meaningful annotations and reproducible benchmarking~\citep{GomezFlores2024BUSBRA:Systems}. It consists of anonymized breast ultrasound images acquired from female patients during routine clinical examinations and includes expert-provided lesion delineations and biopsy-confirmed pathology labels. In addition to image-level diagnostic labels, the dataset incorporates BI-RADS assessments assigned by an experienced ultrasonographer, enabling research across multiple levels of breast cancer risk stratification and lesion characterization. The dataset has been designed to facilitate fair comparison of learning-based methods by providing well-defined partitions and comprehensive annotations commonly required in breast CAD pipelines. We use BUSBRA for a supervised image-level tumor malignancy classification downstream task with two classes: benign (722) and malignant (342). The dataset contains 1,064 images in total. We use 765 images for training, 86 for validation, and 213 for testing.

\textbf{TN5000:} The TN5000 dataset was introduced to enable large-scale research on automated thyroid nodule analysis from ultrasound imaging, with a particular emphasis on malignancy assessment~\citep{Zhang2025TN5000:Classification}. It comprises expertly curated ultrasound images of thyroid nodules collected through a rigorous process of data selection and annotation. Each image is paired with structured annotations following a PASCAL VOC–compatible format, where nodules are explicitly labeled and localized, allowing both region-aware modeling and image-level learning. Diagnostic labels are derived from clinical assessment, and the dataset was intentionally constructed to reduce bias by maintaining a relatively balanced representation of benign and malignant cases at scale. TN5000 has been widely adopted for benchmarking deep learning approaches for thyroid cancer detection due to its size, standardized organization, and clear labeling scheme. We use TN5000 for a supervised image-level thyroid nodule malignancy classification downstream task with two classes: benign and malignant. The dataset contains a total of 5,000 images, including 3,572 malignant nodules and 1,428 benign nodules. We use 3,500 images for training, 500 for validation, and 1,000 for testing. 

\begin{table}[width=.9\linewidth,cols=8,pos=h]
  \caption{Comprehensive statistics of the downstream datasets, including organ, task description, number of classes and data splits.}
  \label{dataset-detailed-table}
  \begin{tabular*}{\tblwidth}{@{}LLLCRRRR@{}}
    \toprule
    Dataset & Organ & Task Description & Classes & Total & Train & Val & Test \\
    \midrule
    TN5000      & Thyroid     & Nodule Malig. & 2 & 5000  & 3500  & 500  & 1000 \\
    FATTY LIV. & Liver       & Fatty Liver Dis. & 2 & 550   & 390   & 50   & 110  \\
    POCUS       & Lung        & Pneumonia, COVID & 3 & 2064  & 1444  & 177  & 443  \\
    BUTTERFLY   & Multi       & Organ Detection & 9 & 41076 & 28053 & 6272 & 6751 \\
    GBCU        & Gallbladder & Lesion Malig. & 3 & 1255  & 1019  & 114  & 122  \\
    AUL         & Liver       & Mass Malig. & 3 & 735   & 529   & 59   & 147  \\
    MMOTU       & Ovary       & Tumor Malig. & 8 & 1469  & 800   & 200  & 469  \\
    BUSBRA      & Breast      & Tumor Malig. & 2 & 1064  & 765   & 86   & 213  \\
    \bottomrule
  \end{tabular*}
\end{table}

\section{US region conditioning (USrc)}
\subsection{Image and USrc Mask Examples}\label{app:C.1}

\begin{figure}[t]
  \vskip 0.2in
  \centering
  \includegraphics[width=1.0\columnwidth]{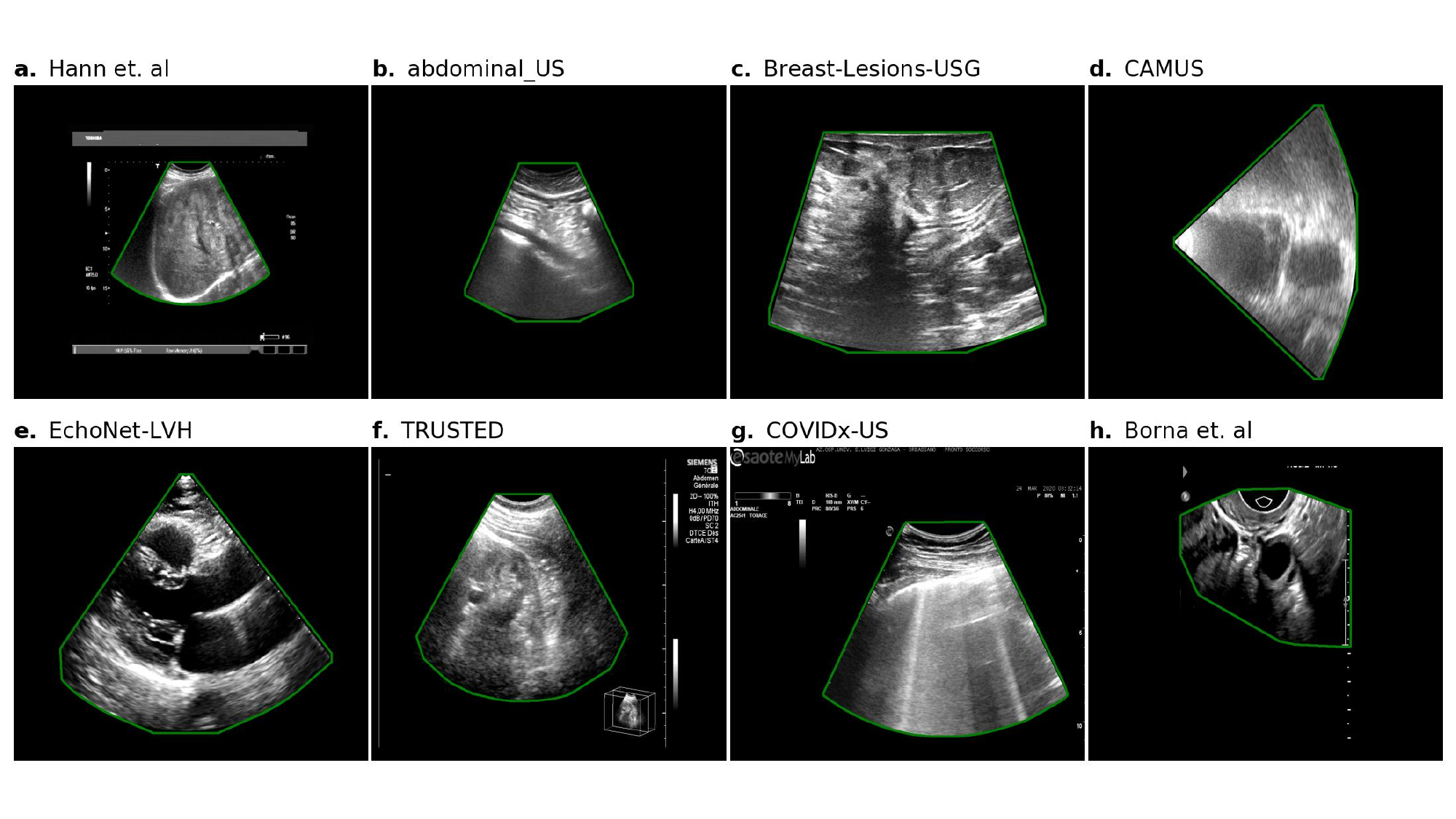}
  \caption{Examples of images and corresponding USrc mask.}
  \label{fig:image_usrc_ex}
\end{figure}

In Fig.~\ref{fig:image_usrc_ex}, we have picked 8 datasets to include examples of images and the corresponding USrc mask. The USrc mask generated by our image processing algorithm is outlined in green and demonstrates the quality of the USrc mask generation.

\subsection{USrc Mask Generation}\label{app:C.2}

To ensure reproducibility, we detail the deterministic image processing pipeline used to generate the binary mask $R$. We first apply a global threshold at a pixel intensity of 15/255, followed by a 5x5 morphological closing to remove gaps in the mask. To adaptively distinguish between different scan types, we identify external contours and rank them by area. We dynamically include the second-largest connected domain only if its area exceeds 50\% of the primary contour. A convex hull transformation is then applied to ensure the mask remains continuous despite internal acoustic shadows, followed by a 25x25 morphological close and a gaussian blur ($\sigma=7$) for final edge smoothing. The USrc mask generation process does not require manual tuning or specialized preprocessing for individual datasets. The same pipeline was applied to 49 of the 50 datasets, with only the prostate US TCIA dataset requiring a prefixed circular geometry.

\section{Training Details}

\subsection{Pretraining Parameters}\label{app:D.1}

We exclusively hold out 5\% of the entire pretraining dataset to use as validation. For our downstream experiments, we evaluate the epoch checkpoint that gets the lowest validation loss during pretraining. All pretraining parameters are detailed in Table~\ref{tab:pretraining_params}.

\begin{table}[width=.7\linewidth,cols=2,pos=h]
  \caption{Shared pre-training hyperparameters for US-JEPA and USrc-JEPA. The models differ only in their use of ultrasound region conditioning.}
  \label{tab:pretraining_params}
  \begin{tabular*}{\tblwidth}{@{}LC@{}}
    \toprule
    \textbf{Configuration} & \textbf{Value} \\
    \midrule
    \textit{Architecture} & \\
    Encoder Backbone & ViT-Base \\
    Predictor Depth & 12 \\
    Predictor Embedding Dim & 384 \\
    Predictor Target Dim & 768 \\
    \midrule
    \textit{Data \& Augmentation} & \\
    Input Size & $224 \times 224$ \\
    Crop Scale & $[0.6, 1.0]$ \\
    Augmentations & H/V Flip, Blur, Artificial Speckle, Contrast \\
    Batch Size & 128 \\
    \midrule
    \textit{Masking (JEPA)} & \\
    Patch Size & 16 \\
    Context Masks (Num, Scale) & 1, $[0.85, 1.0]$ \\
    Target Masks (Num, Scale) & 4, $[0.075, 0.125]$ \\
    Aspect Ratio & $[0.75, 1.5]$ \\
    Min. Patches Kept & 10 \\
    \midrule
    \textit{Optimization} & \\
    Optimizer & AdamW \\
    Total Epochs & 100 \\
    Warmup Epochs & 10 \\
    Base Learning Rate (LR) & $5.0 \times 10^{-5}$ \\
    Start $\rightarrow$ Final LR & $5.0 \times 10^{-6} \rightarrow 5.0 \times 10^{-7}$ \\
    LR Schedule & Cosine Decay with Linear Warmup \\
    Weight Decay $\rightarrow$ Final WD & $0.04 \rightarrow 0.4$ \\
    \bottomrule
  \end{tabular*}
\end{table}

\subsection{Downstream Parameters}\label{app:D.2}

We perform downstream evaluation by training a randomly initialized linear layer on top of the frozen encoder backbone. We utilize early stopping based on the validation loss with a patience of 15 epochs to prevent overfitting. The remaining parameters are detailed in Table~\ref{tab:linear_probing_params}.

\begin{table}[width=.7\linewidth,cols=2,pos=h]
  \caption{Hyperparameters for downstream linear probing on UltraBench datasets.}
  \label{tab:linear_probing_params}
  \begin{tabular*}{\tblwidth}{@{}LC@{}}
    \toprule
    \textbf{Configuration} & \textbf{Value} \\
    \midrule
    Optimizer & AdamW \\
    Base Learning Rate & $1.0 \times 10^{-3}$ \\
    Weight Decay & $1.0 \times 10^{-4}$ \\
    Batch Size & 32 \\
    Input Size & $224 \times 224$ \\
    Loss Function & Cross-Entropy \\
    LR Schedule & Cosine Annealing \\
    Max Training Epochs & 150 \\
    Early Stopping Patience & 15 Epochs \\
    \bottomrule
  \end{tabular*}
\end{table}

\section{Few-Shot Scaling for Linear Probe}

\subsection{Extended Results}\label{app:E.1}
\begin{figure}
  \centering
  \includegraphics[width=\textwidth]{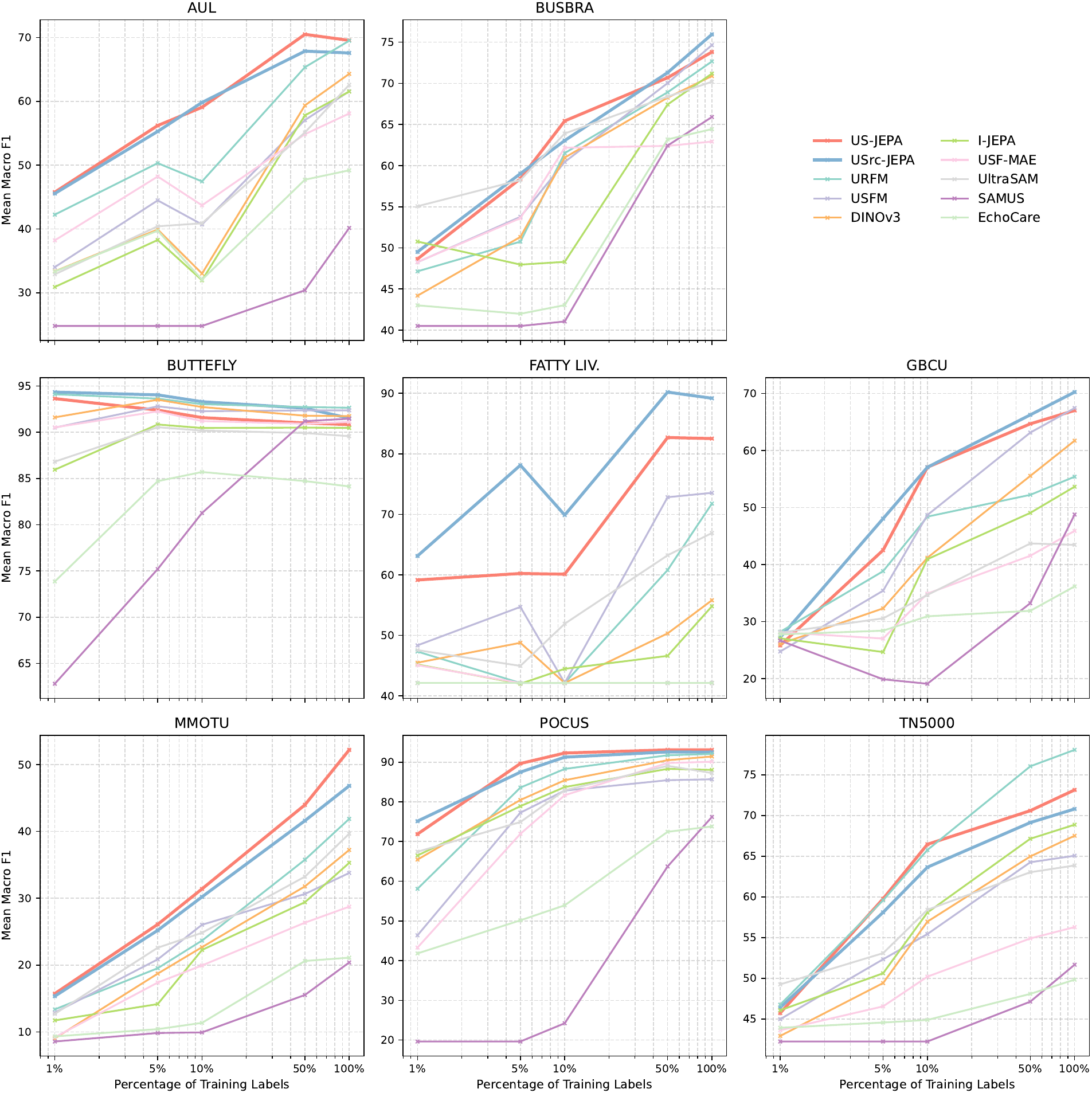}
  \caption{Few-shot classification robustness across all datasets. Here the test set evaluation for linear probe tuned with varying percentages of training data are shown for all models.}
  \label{fig:fewshot_appendix}
\end{figure}

In Fig.~\ref{fig:fewshot_appendix} we include the comprehensive few-shot scaling experiments. For all of our baselines we perform few-shot scaling across five random seeds to see how performance changes on each downstream. In the main paper, we only compare against URFM and USFM  because they were the most competitive baselines.

\section{Robustness to Domain-Specific Corruption Details}

\subsection{Extended Results}\label{app:F.1}

\begin{figure}
  \centering
  \includegraphics[width=\textwidth]{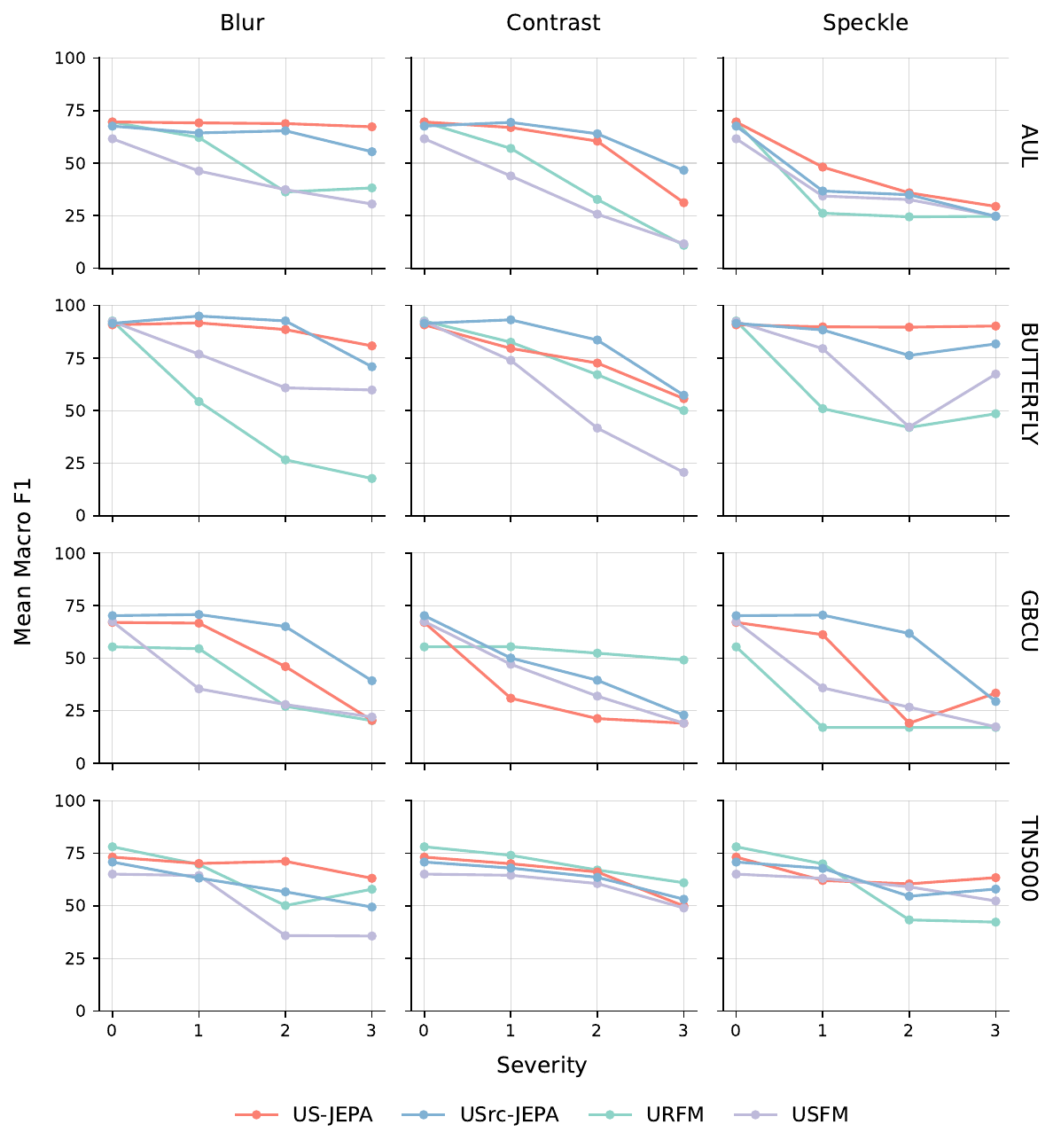}
  \caption{OOD robustness results for remaining 4 downstream datasets.}
  \label{fig:oodrobustness_appendix}
\end{figure}

In Fig.~\ref{fig:oodrobustness_appendix} we include the complete results for the OOD robustness experiment. We show the performance degradation for US-JEPA, USrc-JEPA, URFM and USFM with worsening levels of blur, contrast reduction and artifical speckle corrupting downstream test set images during probe evaluation.

\subsection{Formulas for Corruptions}\label{app:F.2}

To simulate ultrasound imaging artifacts, we define three corruption functions $C(I, \epsilon)$ where $I$ is the input image and $\epsilon$ is the severity level ranging from 1 to 3.

\textbf{Gaussian Blur: } Gaussian blur is modeled as the convolution of the image $I$ with a Gaussian kernel $G$, where the standard deviation $\sigma$ is controlled by the severity $\epsilon$:
\begin{equation}
    I_{blur} = I * G_{\sigma} \quad \text{where } \sigma = \epsilon
\end{equation}
The discrete kernel size for $G$ is adjusted to $2 \lfloor 2\epsilon \rfloor + 1$ to ensure the Gaussian distribution is properly captured.

\textbf{Contrast Depletion: } This corruption shrinks intensities toward the median brightness $\mu_{med}$ of the pixels defined by the USrc mask:
\begin{equation}
    I_{contrast} = \mu_{med} + \alpha(\epsilon) \cdot (I - \mu_{med})
\end{equation}
where the depletion factor $\alpha$ decreases as severity increases ($\alpha \in \{0.7, 0.5, 0.3\}$), effectively making anatomical boundaries less visible.

\textbf{Correlated Speckle Noise: } Speckle is modeled as multiplicative noise. The raw Gaussian noise $\eta$ is spatially correlated via a smoothing kernel $K$ before being applied.
\begin{equation}
    I_{speckle} = I \cdot (1 + \eta_{corr})
\end{equation}
The noise characteristics are functions of severity $\epsilon$:
\begin{itemize}
    \itemsep0em 
    \item Magnitude: $\eta \sim \mathcal{N}(0, \sigma_{noise}^2)$ where $\sigma_{noise} = 0.35\epsilon$.
    \item Correlation: $\eta_{corr} = \eta * K_{size}(\epsilon)$, where the kernel size $K_{size}$ increases with $\epsilon$ to simulate larger speckle grains at higher severity.
\end{itemize}

\subsection{Qualitative Examples}\label{app:F.3}

\begin{figure}
  \centering
  \includegraphics[width=\textwidth]{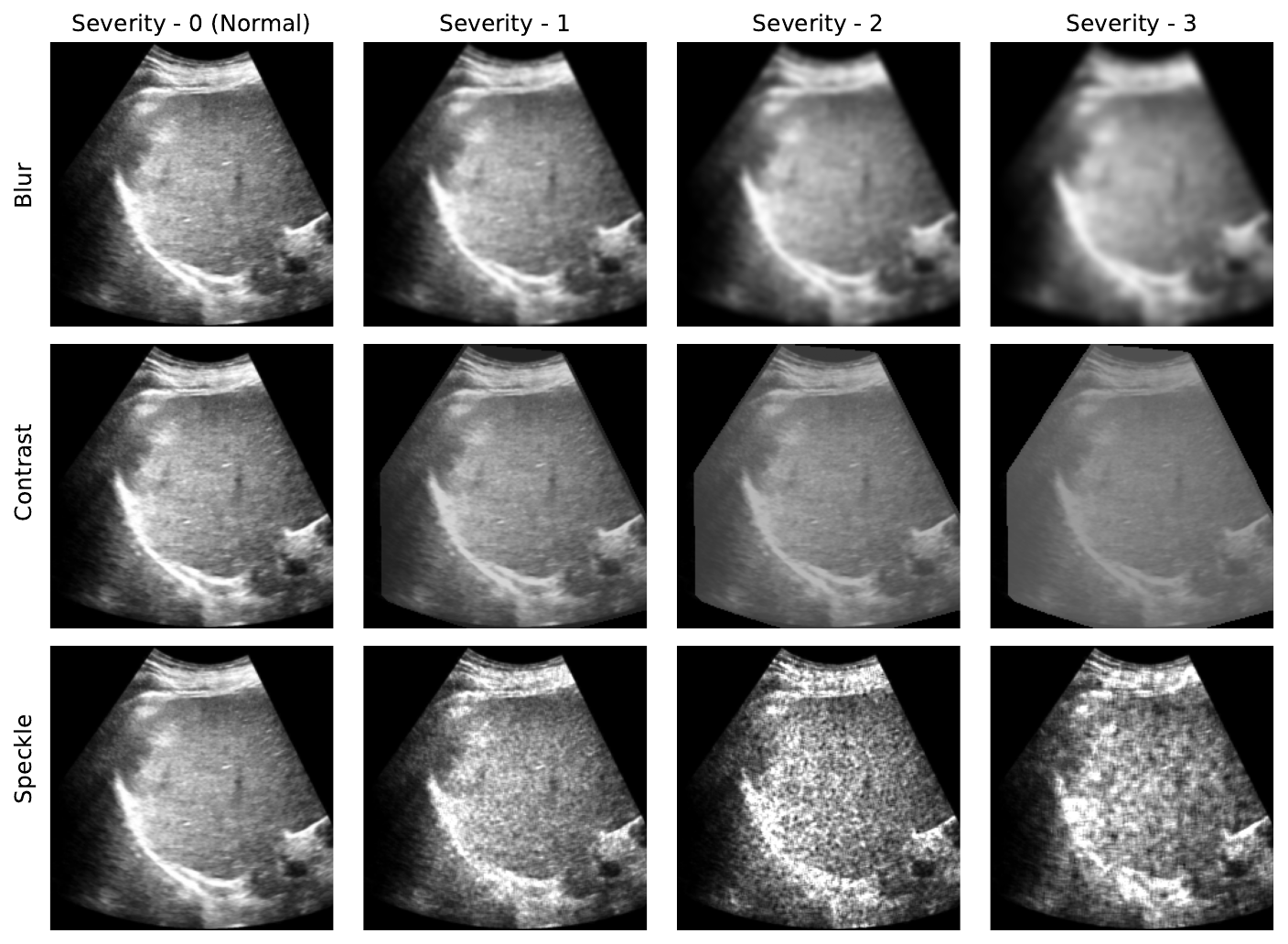}
  \caption{Different corruptions at three severity levels applied to a liver ultrasound from AUL dataset.}
  \label{fig:oodrobustness_examples}
\end{figure}

In Fig.~\ref{fig:oodrobustness_examples}, we have included a qualitative example of the varying severities for each corruption on one example image from the AUL downstream dataset.

\section{Attention Visualizations}\label{app:G}
\begin{figure}
  \centering
  \includegraphics[width=\textwidth]{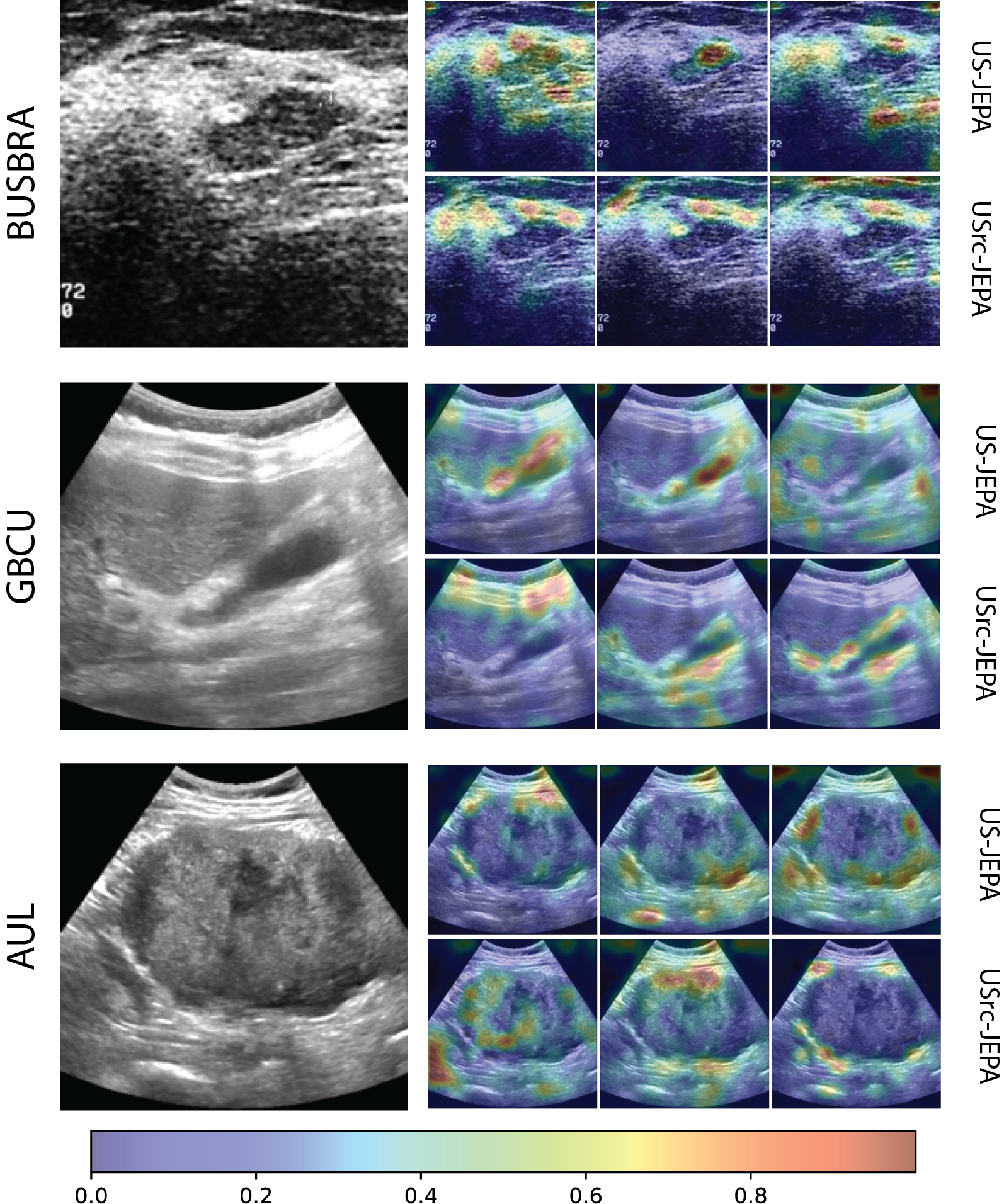}
  \caption{Attention visualization for US-JEPA and USrc-JEPA for three downstream datasets.}
  \label{fig:attviz}
\end{figure}

In Fig.~\ref{fig:attviz}, we show the attention maps from three heads for each US-JEPA and USrc-JEPA across three distinct datasets.

